\pdfoutput=1

\documentclass[11pt]{article}

\usepackage[final]{acl}

\usepackage{times}
\usepackage{latexsym}

\usepackage{booktabs} 
 \usepackage{multirow}
\usepackage{threeparttable}
 
\usepackage[T1]{fontenc}

\usepackage[utf8]{inputenc}

\usepackage{microtype}

\usepackage{inconsolata}

\usepackage{graphicx}
\usepackage{ulem}
\usepackage{float}
\usepackage{xurl}
\usepackage{enumitem}

\usepackage{tcolorbox}
\usepackage{xcolor}
\usepackage{multirow}

\title{Behind Closed Words: Creating and Investigating the forePLay Annotated Dataset for Polish Erotic Discourse \\ 
\textcolor{red}{\small \uline{Warning: This paper contains examples of sexually explicit material}}
} 

\author{Anna Kołos$^{*1}$, Katarzyna Lorenc$^{*1}$, Emilia Wiśnios$^{*2}$, Agnieszka Karlińska$^{1}$\\
$^{1}$ NASK National Research Institute\\
$^{2}$ Independent Researcher \\
$^{*}$ Equal Contribution}

\begin{document}
\maketitle
\begin{abstract}
The surge in online content has created an urgent demand for robust detection systems, especially in non-English contexts where current tools demonstrate significant limitations. We introduce forePLay, a novel Polish-language dataset for erotic content detection, comprising over 24,000 annotated sentences. The dataset features a multidimensional taxonomy that captures ambiguity, violence, and socially unacceptable behaviors. Our comprehensive evaluation demonstrates that specialized Polish language models achieve superior performance compared to multilingual alternatives, with transformer-based architectures showing particular strength in handling imbalanced categories. The dataset and accompanying analysis establish essential frameworks for developing linguistically-aware content moderation systems, while highlighting critical considerations for extending such capabilities to morphologically complex languages.\footnote{Dataset link: \url{https://github.com/ZILiAT-NASK/ForePLay}}
\end{abstract}

\section{Introduction}

The rapid growth of user-generated content online has created an urgent need for effective tools to detect and moderate harmful and inappropriate material. Traditional methods, such as manual review by editors or flagging by users, often fall short due to the sheer volume of content published daily. As a result, there has been an increasing reliance on automated solutions powered by advanced language models and natural language processing techniques.

Securing large language models (LLMs) against the generation of harmful content is another critical concern \cite{yaaseen2024, jialin2024}. While traditional content moderation tools are useful for user-generated content, they are less effective as input/output guardrails for LLMs due to their inability to adapt to new policies and distinguish between user-generated and AI-generated content \cite{llamaguard}. As a result, there is a growing need for datasets and specialized tools that can be integrated with LLMs to prevent the generation of explicit or harmful material, while maintaining their functionality across different applications.

While much of the previous work has focused on detecting toxicity, abusiveness, offensive language, or hate speech \cite{pavlopoulos, gehman, macavaney}, these categories do not encompass the full spectrum of undesired content \cite{markov2023}. Erotic material, in particular, poses a significant risk, especially to minors. Detecting such content is essential not only for creating a safer online environment but also for upholding ethical standards. Although the Digital Services Act (DSA) primarily addresses non-consensual pornography and age-verification requirements, many online platforms, particularly those targeting minors, enforce stricter moderation policies that extend beyond legal obligations. Some platforms may prohibit explicit erotica or any 18+ material, even when it does not violate the law. One application of this research is to support such platforms in enforcing their internal guidelines, which vary depending on the service and its target audience.

Recent advances in deep learning have improved erotic content detection, but progress is hindered by the limited availability of training data, particularly open datasets. Existing tools are primarily designed for English-language content, limiting their effectiveness for other languages. This linguistic bias highlights the need for language-specific datasets and models that can capture the subtle semantic variations inherent in the expression of erotic content in different languages. Moreover, many existing datasets rely on overly simplified binary classification schemes, which fail to capture the complexities of erotic content, further limiting detection systems’ effectiveness in diverse cultural and linguistic contexts.

To address these gaps, we introduce forePLay, the first Polish language manually annotated dataset of erotic content. In contrast to prior efforts, we present a novel multidimensional taxonomy applied to a large Polish language corpus ($n=24,768$), incorporating fine-grained annotations across ambiguity, violence, and social unacceptability dimensions, while ensuring representation of LGBT-specific content from both literary and web-based sources. The corpus is constructed from user-generated content sourced from online fiction repositories and professionally published Polish literary works, offering diverse linguistic and stylistic coverage. The article provides a detailed description of the annotation process and discusses the challenges encountered during annotation, in particular human label variation. A major contribution is also the comprehensive evaluation of erotic content detection models, examining specialized Polish transformer-based models (HerBERT \cite{mroczkowski-etal-2021-herbert} and Polish RoBERTa \cite{polish-roberta}) and Polish-specific LLMs such as the PLLuM family\footnote{For more information, see \url{https://pllum.org.pl/} and \url{https://huggingface.co/CYFRAGOVPL}} and Bielik \cite{Bielik11Bv23a}. We also compare these models with state-of-the-art multilingual and general-purpose models, including both open-source (Mixtral \cite{Jiang2024MixtralOE}, Llama 3.1 \cite{dubey2024llama3herdmodels}, and C4AI Command-R \cite{commandr}) and commercial solutions (GPT-4o \cite{hurst2024gpt}).


\section{Related Work}
\subsection{Existing Datasets}

Prior work has established several benchmark datasets for analyzing textual content with erotic themes, though these exhibit notable limitations in scope and annotation granularity. The most comprehensive corpus, Triplex, comprises $27,000$ literary works totaling $1.62$ billion tokens extracted from Archive of Our Own \cite{achour2016data}, while subsequent contributions include the erotic-books\footnote{\url{https://huggingface.co/datasets/AlekseyKorshuk/erotic-books}} corpus ($n=646$) and the BeaverTails dataset  containing a class of \textit{sexually explicit, adult content} and developed for alignment research \cite{beavertails}. Recent approaches to content analysis have leveraged large language models, as demonstrated in the erotica-analysis dataset\footnote{\url{https://huggingface.co/datasets/openerotica/erotica-analysis}} ($n=15,000$) utilizing GPT-3.5 for automated annotation. Various targeted datasets address binary classification of explicit content, including the Jigsaw corpus and its derivatives ($n=5,100$) \cite{jigsaw-unintended-bias-in-toxicity-classification}, though these typically employ simplified taxonomies. While specialized collections such as the sexting corpus\footnote{\url{https://github.com/mathigatti/sexting-dataset}} ($n=547$) examine specific discourse patterns, existing datasets predominantly focus on English-language content with binary classification schemes. 



\subsection{Erotic Content Detection}

Existing methodological approaches to erotic content detection span a broad spectrum of techniques, though their effectiveness is often constrained by the limitations of available training data. Early work relied on classical machine learning approaches, employing Support Vector Machines (SVM) and Random Forest classifiers trained on social media content \cite{Barrientos2020}, while simpler approaches utilized Naive Bayes classification for web page filtering \cite{4160952}. More recent neural approaches have demonstrated superior performance, as evidenced by Hierarchical Attention Networks applied to historical Latin texts (n = 2,500 sentences), achieving significant improvements over token-based methods \cite{clerice-2024-detecting}. Contemporary research has increasingly leveraged large language models, as demonstrated by CENSORCHAT \cite{DBLP:journals/corr/abs-2403-13250}, which employs knowledge distillation for monitoring dialogue systems. Advanced architectures combining semantic and statistical features through Convolutional Neural Networks (CNN) have shown promise \cite{1598819}. However, these approaches predominantly focus on English language content, limiting their applicability to other languages where linguistic nuances and cultural contexts play crucial roles in content interpretation.

\subsection{Content Moderation Systems}

One key application of erotic content detection datasets is content moderation, covering both user-generated and AI-generated material. Specialized systems like Llama Guard, a Llama2-7b safeguard model developed by Meta, use safety taxonomies to categorize prompts, including sexual content \cite{llamaguard}. However, comparative studies indicate that general-purpose LLMs, such as GPT-4o \citep{hurst2024gpt} and Gemini 1.5 Pro \cite{team2024gemini}, often outperform these systems by achieving a better balance between false positives and false negatives \cite{aldahoul2024}.

Holistic approaches to content moderation, such as those proposed by \citet{markov2023}, combine careful taxonomy design, active learning, and lightweight transformer models. These methods excel in detecting rare harmful content, such as material involving minors. Similarly, the framework introduced by \citet{jialin2024} leverages conceptual features from LLM inference, achieving high accuracy for sexual content with minimal computational cost. However, their findings reveal that LLMs tend to self-censor compared to human-written text. Similar conclusions were drawn by \citet{yaaseen2024}, whose evaluation of OpenAI’s moderation systems highlights GPT models’ fine-tuning to avoid generating sexual content.

These findings highlight the challenges of building scalable, linguistically inclusive moderation frameworks, especially for nuanced categories like erotic material. Most current tools rely on English-centric training and evaluation pipelines \cite{markov2023}. To address cultural and linguistic nuances, particularly in morphologically complex languages like Polish, language-specific detection strategies are essential.

\section{Data}

\subsection{Data Collection}

We constructed a large-scale Polish language corpus ($n = 24{,}768$ sentences) through systematic sampling from two distinct sources: (i) user-generated content from online fiction repositories and (ii) Polish literary works, including translations of world literature. The data collection was carried out by an interdisciplinary team consisting of literary scholars with extensive expertise in contemporary Polish literature, sociologists, NLP specialists, and IT engineers. The corpus comprises demographically diverse material drawn from 905 unique text units and includes a substantial, manually curated representation of LGBTQ+ narratives. This representation was ensured by selecting queer-themed literary texts—primarily original Polish works, supplemented by a smaller number of translated pieces—based on scholarly research and expertise in Polish queer literature. Document-level context is preserved through unique file identifiers. While direct URLs are omitted due to the volatility of online sources, our sampling methodology ensures broad coverage across linguistic registers and social contexts.

Given the potential application of models in supporting moderation, we focused on internet language and amateur writing, which is predominantly anonymous. Consequently, a significant portion of the sentences (69\%) originates from online stories. While the majority of these stories are erotic and non-professional, we included non-erotic stories in a 1:4 ratio to enhance diversity and reduce genre-specific biases. To filter erotic content, we relied on tags and category labels related to erotica provided by the online repositories. All non-professional stories were scraped from publicly available websites, see Appendix \ref{appendix:data-sources} for details. 

To avoid overfitting to specific individual writing styles, we limited the dataset to a maximum of two stories per author.
Additionally, incorporating professional literary works with varying degrees of erotic themes (31\%) aimed to further diversify linguistic patterns. When selecting these texts, availability in digital form was also a key criterion. A total of 22 different literary texts by Polish and international authors were included. This number is significantly lower than the count of unique text units for non-professional content. However, the literary texts were considerably longer, and finding works that met the established criteria proved challenging. Considering the overall proportion of professional works in the corpus, this number should be sufficient.

\subsection{Data Preprocessing}
Our text segmentation pipeline employs the NLTK library \cite{bird-loper-2004-nltk} for sentence boundary detection, which exhibits robust performance despite the inherent variability in web-sourced content quality. To maximize ecological validity, we preserved the original linguistic characteristics of the source material, including non-standard language patterns and orthographic variations, thereby enabling downstream models to generalize effectively to real-world applications. 

The corpus contained a total of 342,546 tokens, as counted using the NLTK tokenizer, with an average sentence length of 13.83 tokens (median 11.0, std 11.53). Detailed token statistics are presented in Table \ref{corpus-stats}.

\begin{table}[ht]
\scriptsize
\centering
\begin{tabular}{l r r r r r}
\toprule
\textbf{Subcorpus} & \textbf{Mean} & \textbf{Std} & \textbf{25\%} & \textbf{50\%} & \textbf{75\%} \\
\midrule
Total & 13.83 & 11.53 & 7.0 & 11.00 & 17.00 \\
Non-professional & 12.43 & 10.32 & 6.0 & 10.00 & 16.00 \\
Professional & 16.90 & 13.31 & 9.0 & 14.00 & 21.00 \\
\bottomrule
\end{tabular}
\caption{Token Count Statistics}
\label{corpus-stats}
\end{table}

Inconsistencies in text segmentation and discrepancies in automatically detected sentences arose from the fact that non-professional writers often did not use proper punctuation, making it challenging for the NLTK library to accurately identify sentence boundaries.

\section{Annotation Process}

To assemble the annotation team, three female annotators were allocated from our internal organization. To ensure gender balance, three part-time male annotators were recruited as external contractors. All selected annotators underwent training and guidance to maintain annotation consistency and quality throughout the project.

This process resulted in a gender-balanced team of six annotators (3 male, 3 female), aged 20–40. We included annotators with backgrounds in linguistics, literature, or related fields to ensure sensitivity to language nuances. To minimize potential bias, the annotators were not provided with metadata that could link specific samples to genre types. Furthermore, samples from each genre type were distributed evenly among all annotators. Each sentence was independently assessed by three annotators. Final labels were determined by majority vote. In cases where all three labels differed ($830$ out of $24,768$ samples, representing $3.35\%$ of the total observations), a superannotator made the final decision. The superannotator, an NLP specialist and senior member of the annotation team, was responsible for overseeing consistency and quality throughout the annotation process. This role encompassed resolving disagreements among annotators, providing additional guidance, and conducting spot checks to ensure adherence to the defined guidelines.

\subsection{Annotation Scheme}
Based on the expert knowledge of the content, a set of five possible exclusive labels to annotate the samples was designed. This included the following categories: \textit{erotic} (e), \textit{ambiguous} (a), \textit{violence-related} (v), \textit{socially unacceptable behaviors} (u), and \textit{neutral} (n). Some examples of the annotated text and its English translation are presented in Table \ref{table:examples}.

\begin{table*}[t]
\centering
\small
\renewcommand{\arraystretch}{1.2}
\begin{tabular}{p{6.5cm}p{6.5cm}p{2cm}}
\toprule
\textbf{Original sentence in Polish} & \textbf{English translation} & \textbf{Label} \\
\midrule
``Mój mężczyzna...''--- usłyszałem szept i zanim się zorientowałem zostałem bez koszulki, a Ty siedziałaś naprzeciw mnie, całując zachłannie moje usta i błądząc dłońmi po torsie, lekko go drapiąc. & ``My man...''--- I heard a whisper and before I knew it I was left shirtless and you were sitting across from me, kissing my lips greedily and wandering your hands over my torso, lightly scratching it. & erotic \\
\midrule
Był cały twardy. & He was all hard. & ambiguous \\
\midrule
Ten gad ją zdradzał! & This rat was cheating on her! & neutral \\
\bottomrule
\end{tabular}
\caption{Examples of annotations in the dataset}
\label{table:examples}
\end{table*}

Sentences labeled as \textit{erotic} typically describe sexual activities and desires, advanced flirting with evident erotic undertones, references to past sexual activities, or explicit sexual fantasies. However, it is important to note that mere mention of genital terms or publicly accepted romantic behaviors, such as kissing or holding hands, does not warrant classification as sexual. Similarly, descriptions of physiological processes or neutral discussions of erotic or sexual topics (e.g., from the psychological perspective) are excluded from this category and considered neutral.

The \textit{violence-related} category is reserved exclusively for sentences that include explicit sexual harassment, rape, lack of consent, or any other sort of non-consensual violence which concerns sexual intentions or activities. It is essential to distinguish consensual BDSM behaviors, as well as non-sexual acts of violence (based solely on the sentence-level analysis), as they do not fall under this category.

The category of \textit{socially unacceptable behaviors} includes sentences that describe sexual behaviors considered illegal, generally taboo, or violative of social norms, such as zoophilia, necrophilia, pedophilia, incest, and other sexual deviations. This category takes precedence over \textit{violence-related} label, meaning that if a sentence could be classified as both \textit{socially unacceptable} and \textit{violence-related}, it should be categorized under \textit{socially unacceptable behaviors}.

Another label covers context-related sentences, which fall under the category of \textit{ambiguous} samples. These sentences are identified based on the conviction that—given commonly recognized patterns of sexual descriptions—they evoke erotic or sexual connotations. However, in a neutral context, they could be interpreted as describing non-sexual behaviors or acts.

Lastly, sentences that did not fall into any of the above-mentioned categories were labeled as \textit{neutral}. On one hand, these typically include non-sexual content on any topic. On the other, this category also comprises general statements about human sexuality or intimate parts that do not typically evoke erotic associations. While such sentences may be sex-related, they are not defined as erotic. 

The annotation guidelines, developed by literary scholars and linguists based on domain expertise and material analysis, are described in detail in Appendix~\ref{appendix:guidelines}.

Following the completion of the superannotation process, the final dataset consists of two primary classes: 68.14\% of the samples are labeled as \textit{neutral}, and 25.68\% as \textit{erotic}. The remaining categories are minor, with 5.43\% labeled as \textit{ambiguous}, 0.47\% as \textit{socially unacceptable behaviors}, and 0.28\% as \textit{violence-related}. Distribution of labels across the subcorpora are presented in Table \ref{tab:label-stats}.

The percentage of violence-related examples, as well as those representing socially unacceptable behaviors, was scarce, which was expected, as the data collection process did not specifically target texts exploring such themes. Additionally, while only 15\% of the texts in the dataset were sourced from non-erotic stories, the randomization process employed to minimize bias resulted in the inclusion of neutral sentences even from texts classified as erotic, as these are not exclusive to non-erotic stories.

\begin{table}[t]
\tiny
\centering
\begin{tabular}{l r r r r r}
\toprule
\multirow{2}{*}{\textbf{Subcorpus}} & \multicolumn{5}{c}{\textbf{Label Categories}} \\
\cmidrule(lr){2-6}
& Erotic & Ambiguous & Violence & Unacceptable & Neutral \\
\midrule
Total & 6,361 & 1,344 & 69 & 116 & 16,878 \\
Non-professional & 2,937 & 1,277 & 52 & 95 & 12,649 \\
Professional & 3,424 & 67 & 17 & 21 & 4,229 \\
\bottomrule
\end{tabular}
\caption{Distribution of Label Categories Across Dataset Subcorpora}
\label{tab:label-stats}
\end{table}



\subsection{Annotation Quality}

The process of quality evaluation revealed various challenges faced by the annotators, highlighting the demanding nature of the task. The designed labels were intended to be applied exclusively, meaning that each sample could only be categorized as either neutral, sex-related, or ambiguous. Within the sex-related category, overlaps were mitigated by a clear hierarchical structure. As previously mentioned, in cases where a sample exhibited characteristics of both violence and socially unacceptable behaviors, priority was given to the latter. All the other sex-related samples contributed to the erotic class. Due to the distinctiveness of rare cases of sexual violence and non-acceptable sexual behaviors, this aspect did not pose significant concerns. Annotators applied these two labels only incidentally, with individual frequencies ranging from 0.2\% to 0.6\%.

Firstly, the annotation task itself—particularly given the wide range of non-normative language patterns found in non-professional short stories—was inherently subjective. Secondly, the prevalence of misspellings, controversial vocabulary, and grammatical errors, which diminished the clarity of the authors' intentions and made rational decision-making exceedingly difficult. Thirdly, the category of erotic sentences turned out to be extremely broad, spanning from evident and explicit samples, often including vulgar language, to much subtler erotic tones. Consequently, the \textit{ambiguous} label, designed to account for context-related uncertainty, offered a means of resolving indecision between \textit{sexual} and \textit{neutral} classifications. This created a potential risk of overusing the ambiguous label as a way to mitigate uncertainty in borderline cases. 

Human label variation (HLV) \cite{plank2022}, inherently present in our task of annotating erotic sentences, raised several questions regarding the dataset design (see the Section \ref{sec:discussion}). To preserve the authenticity of the annotation process, we chose not to intervene beyond the initial training phase. For the dataset's initial release, we opted to follow the conventional practice of data aggregation featuring a ground truth. 

To guarantee annotation reliability, we calculated Cohen's Kappa for each pair of annotators across all categories. Furthermore, we evaluated inter-annotator agreement for each category individually. Detailed results are presented in Table~\ref{table:annotations}.

Agreement in most pairs was consistent, ranging from 0.66 to 0.71, regardless of annotators' gender. However, pairs involving annotator Fem1 presented a notable deviation, as Fem1 emerged as an outlier, largely due to an excessive use of the ambiguous label.

In addition to pairwise agreement, the overall consistency of the entire sample was assessed using Krippendorff's Alpha. The value was quite low (0.387). However, after excluding the Fem1 annotation, which significantly deviated from the other annotators, the value increased to 0.716, indicating a satisfactory level of agreement in the annotation process. 


\begin{table}
\scriptsize
\renewcommand{\arraystretch}{1.5} 
\centering
\begin{tabular}{c|cccccc}
\toprule
\textbf{Annotators} & \textbf{all} & \textbf{a} & \textbf{e} & \textbf{u} & \textbf{v}  \\
\midrule
\textcolor{teal}{Fem1} / \textcolor{olive}{M1} & 0.1750 & 0.0532 & 0.3602 & 0.2545 & 0.1984 \\
\textcolor{teal}{Fem1}  / \textcolor{teal}{Fem2}  & 0.1808 & 0.0370 & 0.3193 & 0.3952 & 0.2911  \\
\textcolor{teal}{Fem1} / \textcolor{teal}{Fem3} & 0.1415 & 0.0151 & 0.2890 & 0.2740 & 0.2271  \\
\textcolor{olive}{M1} / \textcolor{teal}{Fem2} & 0.7044 & 0.3505 & 0.7513 & 0.3728 & 0.4602  \\
\textcolor{teal}{Fem2} / \textcolor{olive}{M2} & 0.6910 & 0.0833 & 0.7149 & 0.5074 & 0.1696  \\
\textcolor{teal}{Fem2} / \textcolor{olive}{M3} & 0.6610 & 0.0037 & 0.6923 & 0.3973 & 0.3237  \\
\textcolor{teal}{Fem2} / \textcolor{teal}{Fem3} & 0.7187 & 0.2683 & 0.7728 & 0.6718 & 0.4078  \\
\textcolor{olive}{M2} / \textcolor{olive}{M3} & 0.7030 & 0.0368 & 0.7261 & 0.4191 & 0.2481  \\
\bottomrule
\end{tabular}
\caption{Cohen's Kappa values for each annotator pair across all categories and for each individual category, broken down by gender (Fem -- female, M -- male).} 
\label{table:annotations}
\end{table}

\section{Experiments}

We utilize a diverse set of language models for erotic content detection, ranging from specialized Polish encoder-based models and developed Polish-specific LLMs to state-of-the-art commercial LLMs. This variety enables a comprehensive comparison across different model architectures and training methodologies. Additionally, the translated English prompt for the LLMs is included in the appendix on Figure~\ref{fig:prompt-template-en}. 


\subsection{Datasets}
The original dataset consists of five distinct label categories, but due to significant class imbalance, additional experiments were conducted with modified label groupings as outlined below:
\begin{itemize}[itemsep=0pt,parsep=0pt,topsep=0pt]
\item \textbf{Basic}: A binary classification setup where the dataset was reduced to two classes: \textit{neutral} and \textit{erotic} labels.
\item \textbf{Core}: A three-class classification setup where the dataset included \textit{neutral}, \textit{erotic}, and \textit{ambiguous} labels.
\item \textbf{Extended}: A four-class classification setup that expanded the dataset to include \textit{neutral}, \textit{erotic} and \textit{ambiguous} labels, with a merged category combining \textit{violence-related} and \textit{socially unacceptable behaviors}.
\item \textbf{Full}: The full dataset containing all five original categories.
\end{itemize}

\subsection{Specialized Polish Transformer-based Models}
Encoder-based models for Polish were selected based on the KLEJ Benchmark leaderboard~\citep{rybak2020klej}. Two top-performing models across multiple downstream tasks were chosen: HerBERT and Polish RoBERTa. HerBERT is a Polish-language adaptation of BERT, trained on a Polish corpus, while Polish RoBERTa is an optimized version featuring a unigram tokenizer, whole-word masking, and an expanded vocabulary of 128k entries. In our experiments, both models were fine-tuned on each dataset configuration (\textbf{Basic}, \textbf{Core}, \textbf{Extended}, \textbf{Full}).

\subsection{Polish Large Language Models}
Our evaluation framework encompasses three variants from the PLLuM (Polish Large Language Model) family: \textbf{Llama-3.1-8B-PLLuM}, \textbf{PLLuM-Mistral-12B}, and \textbf{PLLuM-Mixtral-8x7B}, adapted from their respective base architectures. These models underwent pre-training on 147B Polish language tokens and subsequent instruction tuning. We additionally employed \textbf{Bielik 2.3}, trained on 200B tokens and aligned using DPO-positive on 66,000 examples. To assess adaptation capabilities, we fine-tuned Llama-3.1-8B-PLLuM and PLLuM-Mistral-12B on our classification task. All base models were evaluated under 0-shot, 1-shot, and 5-shot configurations, while fine-tuned variants underwent standard supervised evaluation.

\subsection{State of the Art Large Language Models}

For comparison with LLMs fine-tuned strictly for Polish, we employed a selection of state-of-the-art multilingual and general-purpose models, including both open-sourced and commercial solutions. Among these, \textbf{Mixtral}, characterized by its modular and scalable architecture, was utilized in both the 8x22B and 8x7B configurations, alongside \textbf{Mistral 12B}, designed as a lightweight alternative. We also tested \textbf{Llama 3.1 70B-Instruct} and its smaller counterpart \textbf{Llama 3.1 8B-Instruct}, both fine-tuned for instruction-following tasks, as well as \textbf{C4AI Command-R}, developed for command-following applications. Additionally, we incorporated \textbf{GPT-4o}, frequently used for tasks requiring generalization across various domains.

\section{Results}

Table~\ref{table:results_bert} highlights the macro-F1 performance metrics of the HerBERT and RoBERTa models across classification tasks involving different dataset configurations. In binary classification scenarios, both models exhibit robust performance. However, as the number of classes increases, a decline in macro-F1 scores is observed. Notably, RoBERTa, particularly in its Base configuration, demonstrates superior performance relative to HerBERT, with pronounced advantages evident in tasks on the \textbf{Basic} and \textbf{Extended} datasets.

\begin{table}[!hbp]
\tiny  
\renewcommand{\arraystretch}{1.5} 
\centering
\begin{tabular}{c|cc|cc}
\toprule
\multicolumn{1}{c|}{\multirow{2}{*}{\textbf{Dataset}}} & \multicolumn{2}{c|}{\textbf{HerBERT}} & \multicolumn{2}{c}{\textbf{RoBERTa}} \\
\cmidrule{2-5}
 & \textbf{Base} & \textbf{Large} & \textbf{Base} & \textbf{Large} \\
\midrule
Basic & 0.929 & 0.939 & \textbf{0.944} & 0.943 \\
Core & 0.702 & 0.738 & 0.738 & \textbf{0.748} \\
Extended & 0.693 & \textbf{0.746} & 0.704 & 0.734 \\
Full & 0.632 & 0.648 & \textbf{0.707} & 0.664 \\
\bottomrule
\end{tabular}
\caption{Macro-F1 scores for HerBERT and RoBERTa models across different dataset versions.}
\label{table:results_bert}
\end{table}

\begin{table*}[h]
\scriptsize
\renewcommand{\arraystretch}{1.5} 
\centering
\begin{tabular}{c|ccc|ccc|ccc|ccc}
\toprule
\textbf{Model} & \multicolumn{3}{c|}{\textbf{Basic}} & \multicolumn{3}{c|}{\textbf{Core}} & \multicolumn{3}{c|}{\textbf{Extended}} & \multicolumn{3}{c}{\textbf{Full}} \\
 & 0-shot & 1-shot & 5-shot & 0-shot & 1-shot & 5-shot & 0-shot & 1-shot & 5-shot & 0-shot & 1-shot & 5-shot \\
\midrule
GPT-4o                  & 0.888 & 0.873 & 0.891 & 0.640 & 0.605 & 0.618 & 0.425 & 0.410 & 0.415 & 0.340 & 0.334 & 0.349 \\
C4AI Command-R               & 0.684 & 0.730 & 0.714 & 0.420 & 0.486 & 0.454 & 0.322 & 0.374 & 0.363 & 0.257 & 0.292 & 0.299 \\
Llama 3.1 8B-Instruct &    0.789   &  0.736    & 0.709     &   0.562  &     0.517   &   0.514  &   0.388   &      0.364 & 0.378    &  0.325    &   0.283   & 0.316 \\
Llama 3.1 70B-Instruct                    & 0.837 & 0.845 & 0.846 & 0.595 & 0.641 & 0.630 & 0.396 & 0.417 & 0.408 & 0.329 & 0.344 & 0.364 \\
Mistral 12B &   0.872   &   0.847   &  0.862    &   0.58  &   0.579   &  0.610   &    0.423    &   0.411   & 0.42    & 0.354     &   0.349   & 0.338 \\
Mixtral 8x7B &  0.867 &0.87    &  0.862   &  0.616   &   0.622   &  0.621   &   0.440     &   0.451   &  0.435   &  0.364    &    0.364  & 0.348 \\
Mixtral 8x22B            & 0.872 & 0.854 & 0.833 & 0.633 & 0.642 & 0.610 & 0.424 & 0.440 & 0.406 & 0.335 & 0.361 & 0.344 \\
Bielik-11B-v2.3-Instruct              & 0.848 & 0.858 & 0.868 & 0.601 & 0.597 & 0.607 & 0.427 & 0.416 & 0.423 & 0.312 & 0.437 & 0.48 \\
\midrule
PLLuM-Mistral-12B         & 0.894 & 0.842 & 0.844 & 0.656 & 0.610 & 0.613 & 0.456 & 0.441 & 0.427 & 0.401 & 0.359 & 0.401 \\
PLLuM-Mixtral-8x7B        & 0.874 & 0.902 & 0.921 & 0.647 & 0.662 & 0.670 & 0.452 & 0.456 & 0.426 & 0.422 & 0.406 & 0.422 \\
Llama-3.1-8B-PLLuM        & 0.641     & 0.737     & 0.709     & 0.425     & 0.446     & 0.425    & 0.301     & 0.324     & 0.326     & 0.314     &  0.274    & 0.273      \\
\midrule
PLLuM-Mistral-12B (SFT)     &  & 0.946    &  & & \textbf{0.792}  &  & & 0.458   &  &  & \textbf{0.488}  & \\ 
Llama-3.1-8B-PLLuM (SFT)     &      & \textbf{0.947}     &      &      & 0.764     &      &      & \textbf{0.461}    &      &      & 0.435         & \\
\bottomrule
\end{tabular}
\caption{Macro-F1 scores for various models across different dataset configurations and shot numbers.}
\label{table:results_llm}
\end{table*}

In the case of LLMs, models from the PLLuM family consistently demonstrate strong performance across varying label counts and shot configurations, with particular emphasis on the PLLuM-Mixtral-8x7B, which achieves outstanding macro-F1 scores across a wide range of label counts, notably reaching 0.921 for \textbf{Basic} data and 0.670 for \textbf{Core} data in the 5-shot setting. Additionally Llama-3.1-8B-PLLuM (SFT) and PLLuM-Mistral-12B (SFT) show significant improvements when subjected to supervised fine-tuning, resulting in enhanced performance, with the highest macro-F1 scores observed across various label counts.

Fine-tuned LLMs exhibit performance levels comparable to traditional transformer-based architectures when evaluated on datasets labeled as \textbf{Basic} and \textbf{Core}, where all categories are sufficiently represented. This suggests that in scenarios characterized by balanced class distributions, where each class is supported by an adequate number of training examples, the performance differential between these methodological approaches is minimal. Conversely, for datasets labeled as \textbf{Extended} and \textbf{Core}, which are marked by pronounced class imbalance, models built upon the BERT architecture begin to demonstrate a distinct advantage. These models excel in effectively addressing the challenges posed by imbalanced data distributions, exhibiting superior capability in identifying underrepresented categories. This highlights their enhanced adaptability and robustness in handling more complex scenarios involving skewed training data distributions.

In the context of multilingual models trained with a primary emphasis on the English language, the results are generally inferior compared to models specifically adapted for the Polish language. The performance varies depending on the configuration, including the number of labels and the number of shots, with different models excelling under different conditions. However, overall, the Mixtrals and GPT-4o models consistently yielded the most favorable results.

The full results of all evaluated models—including the baseline models, the ones adapted to the Polish language from the PLLuM family, and the ones further fine-tuned on our dataset—are presented in Table~\ref{table:results_llm}. 
A detailed account of refusals and responses that could not be classified into any of the predefined categories is presented in Table~\ref{table:refusal_llm}. Among the evaluated models, Llama 3.1 8B-Instruct recorded the highest number of refusals, reaching a total of 848. In contrast, models that consistently provided clear and categorizable outputs included GPT-4o and Mixtral 8x22B, as well as those additionally fine-tuned on our dataset.


\begin{table*}[!htbp]
\scriptsize
\renewcommand{\arraystretch}{1.5} 
\centering
\begin{tabular}{c|ccc|ccc|ccc|ccc|c}
\toprule
\textbf{Model} & \multicolumn{3}{c|}{\textbf{Basic}} & \multicolumn{3}{c|}{\textbf{Core}} & \multicolumn{3}{c|}{\textbf{Extended}} & \multicolumn{3}{c|}{\textbf{Full}} & \textbf{Total} \\
 & 0-shot & 1-shot & 5-shot & 0-shot & 1-shot & 5-shot & 0-shot & 1-shot & 5-shot & 0-shot & 1-shot & 5-shot & \\
\midrule
GPT-4o                  & 0     & 0     & 0     & 0     & 0     & 0     & 0     & 0     & 0     & 0     & 0     & 0     &  \textbf{0}  \\
C4AI Command-R               & 9     & 2     & 14    & 9     & 5     & 10    & 7     & 1     & 9     & 4     & 1     & 14    & 95 \\
Llama 3.1 8B-Instruct &    42  &   150   &    12  &  38   &   99   &  14   &  134   &   62   &   8   &  66   &  208    &   15   &  848\\
Llama 3.1 70B-Instruct & 8     & 0     & 0     & 33    & 0     & 0     & 46    & 0     & 0     & 49    & 0     & 0     & 136 \\
Mistral 12B&    0  &   0   & 0     &   0  &   0   &  0   &   0  &    0  &  1    &  0   &   0   &   0   &  1 \\
Mixtral 8x7B&   0   &   0   &  2    &  1   &   0   &   2  &   1  &  5    &    7  & 3    &   0   &    4  &  25 \\
Mixtral 8x22B            & 0     & 0     & 0     & 0     & 0     & 0     & 0     & 0     & 0     & 0     & 0     & 0     &  \textbf{0}  \\\
Bielik-11B-v2.3-Instruct              & 116   & 1     & 0     & 115   & 2     & 0     & 148   & 1     & 3     & 2     & 5     & 1     & 394 \\
\midrule
PLLuM-Mistral-12B         & 3     & 5     & 1     & 3     & 2     & 3     & 8     & 7     & 0     & 6     & 10    & 5     & 53 \\
PLLuM-Mixtral-8x7B        & 0     & 0     & 1     & 0     & 0     & 2     & 0     & 0     & 2     & 0     & 1     & 2     & 8 \\
Llama-3.1-8B-PLLuM        & 1     & 4     & 0     & 1     & 4     & 3     & 1     & 3     & 5     & 0     & 4     & 1     & 27 \\
\midrule
PLLuM-Mistral-12B (SFT)     &      & 0     &      &      & 0     &      &      & 0     &      &      & 0     &      & \textbf{0} \\
Llama-3.1-8B-PLLuM (SFT)     &      & 0     &      &      & 0     &      &      & 0     &      &      & 0     &      & \textbf{0} \\
\bottomrule
\end{tabular}
\caption{Number of unidentified responses or refusals.}
\label{table:refusal_llm}
\end{table*}

\section{Discussion}\label{sec:discussion}
The task of multi-dimensional eroticity annotation is highly complex and inherently subjective, necessitating further and more advanced reflection on the Human-Language Variation (HLV) framework, which has become an increasingly important open problem in recent years \cite{plank2022, uma, pavlick2019}. The researchers highlight the fact that disagreements in the annotation process should not be dismissed as mere \textit{noise} in the search for a single ground truth. On the contrary, such disagreements can often reflect plausible alternative judgments, suggesting that these annotations may not always be categorical \cite{marneffe}. However, as demonstrated by our experience evaluating annotation quality, distinguishing between clear errors—where no plausible argument could justify the choice of a particular label—and alternative judgments is a critical issue that warrants deeper concern and further discussion. Within the context of erotic content detection, we believe it is valuable to advance current research by utilizing datasets with multiple plausible labels and exploring the approach of Learning with Disagreements (LeWiDi) \cite{uma} or Multiple Ground Truth \cite{anand2024}. This approach could offer more nuanced insights into the field by comparing the performance of models trained on aggregated annotations versus those trained on multiple possible labels.

Our experimental results demonstrate clear performance patterns across model architectures and linguistic contexts, with specialized Polish language models consistently outperforming multilingual alternatives in erotic content detection. The encoder-based architectures, particularly RoBERTa, achieve robust macro-F1 scores in binary classification ($0.929-0.944$), though performance degrades with increasing categorical complexity (dropping to $\sim0.66$ for five-class classification). This degradation pattern reveals fundamental challenges in maintaining discriminative power across finer-grained categorical distinctions, particularly in cases requiring subtle cultural and contextual understanding.

The PLLuM family of models exhibits exceptional few-shot learning capabilities, with PLLuM-Mixtral-8x7B achieving macro-F1 scores of $0.921$ in 5-shot settings for binary classification, while supervised fine-tuning of PLLuM-Mistral-12B yields the highest overall performance (macro-F1 = $0.946$). However, the diminishing returns observed with increasing model size, coupled with the persistent challenge of handling ambiguous content in imbalanced datasets, suggests that architectural sophistication alone may not overcome the fundamental challenges of erotic content detection in morphologically complex languages. These findings underscore the importance of language-specific model development for sensitive content detection tasks, rather than relying solely on increasingly large multilingual models.

\begin{figure}
    \centering
    \includegraphics[width=\columnwidth]{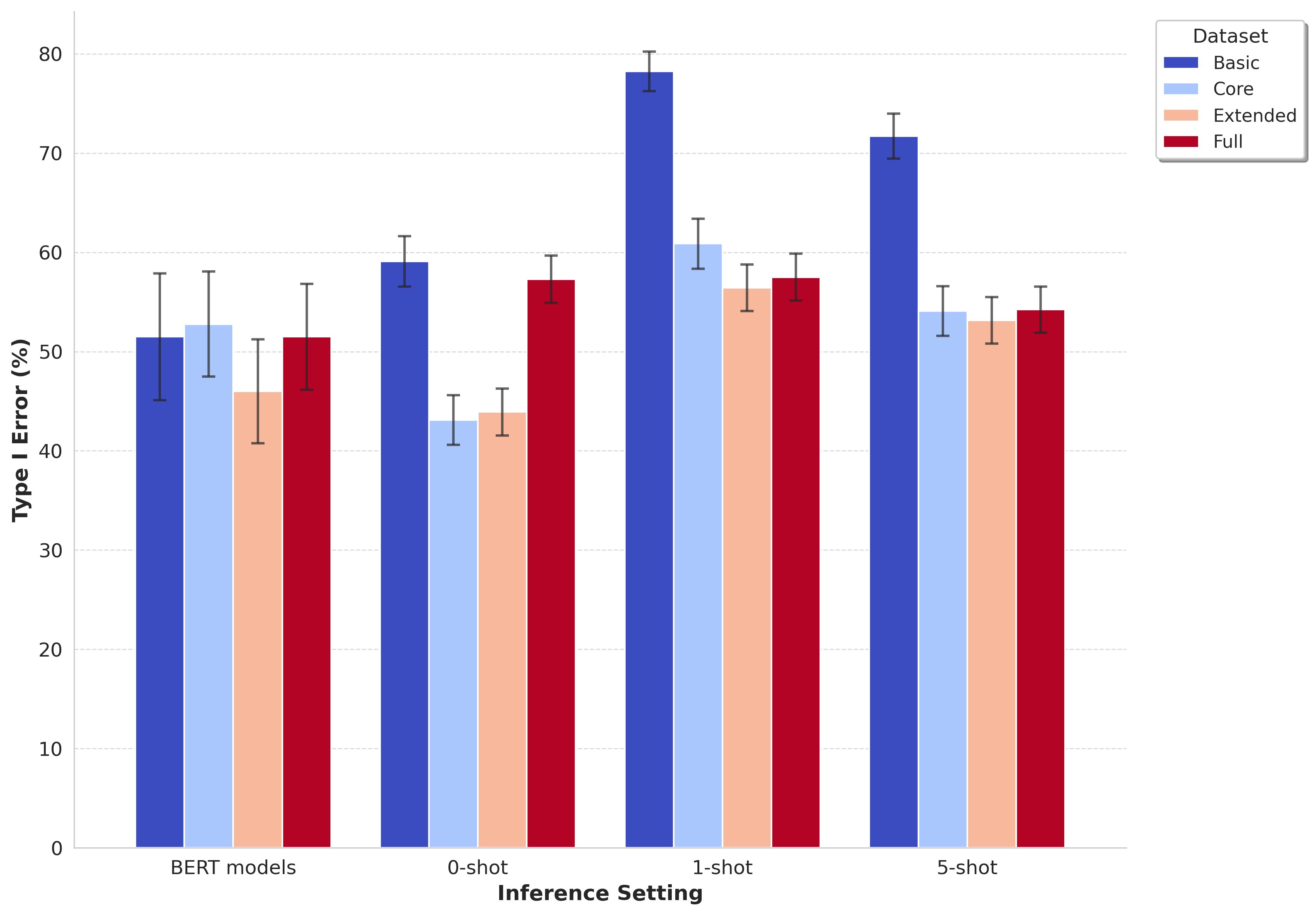}
    \caption{Type I errors percentage across datasets and inference settings. Error bars represent 95\% confidence interval.}
    \label{fig:typeIerror}
\end{figure}

Beyond overall performance metrics, understanding error patterns is crucial for content moderation deployment, where both false positives (over-censorship) and false negatives (missed inappropriate content) carry significant consequences. We calculated Type I error percentages as the number of false positives divided by the sum of false positives and false negatives to assess model precision across different classification scenarios. Full results regarding Type I and Type II errors for each model are presented in the Appendix \ref{appendix:errors}. The analysis reveals that Type I error rates decrease substantially with classification complexity—from approximately 60-80\% in binary classification (Basic) to 40-50\% in the full five-class scenario (Figure \ref{fig:typeIerror}). This pattern indicates higher Type I error rates in simpler tasks and higher Type II rates in complex multi-class scenarios, highlighting the fundamental trade-off between precision and sensitivity in erotic content detection tasks. 

\section{Conclusions}
This work presents forePLay, the first comprehensive Polish language dataset for erotic content detection, introducing a novel multidimensional taxonomy encompassing ambiguity, violence, and social acceptability dimensions. Through systematic empirical evaluation across multiple model architectures, our analysis demonstrates the superiority of specialized Polish language models over multilingual alternatives, particularly in handling nuanced categorical distinctions. The performance patterns observed in transformer-based architectures and few-shot learning scenarios provide valuable insights for developing language-specific content moderation systems. Our findings underscore the critical importance of culturally and linguistically adapted approaches in content detection tasks, while the presented dataset and evaluation framework establish essential groundwork for future research in multilingual content moderation systems, particularly for morphologically complex languages.



\section{Data Availability}

A subset of erotic and ambiguous sentences, totaling $3{,}704$ samples, has been released on our public Github account. Due to ethical considerations, we opted to exclude the minor harmful classes of \textit{violence-related} content and \textit{socially unacceptable behaviors}. ForePLay Dataset Release 1.0 consists of $2{,}728$ sentences labeled as \textit{erotic}, and $976$ sentences labeled as \textit{ambiguous}, which corresponds to 43\% and 73\% of the original dataset, respectively. Neutral sentences were not included in this release, as such data is readily obtainable from publicly available corpora and may be easily used for comparative purposes in downstream tasks.
This subset has undergone additional copyright verification and was made available following legal consultations. The released data and its license comply with new legal regulations that came into effect after the data collection and annotation process had been completed.

\section*{Limitations}

Our study reveals several important limitations that warrant discussion. While our dataset represents a significant contribution to Polish language resources, it exhibits sampling biases through its predominance of content from online fiction repositories ($69\%$). This may not fully capture the linguistic diversity of erotic discourse across different domains and registers. The relative underrepresentation of professional literary works ($31\%$) potentially limits our understanding of more sophisticated expressions of erotic content.

A significant methodological limitation emerges from our binary approach to violence and socially unacceptable behaviors. The current taxonomy, while practical for annotation purposes, may oversimplify the complex spectrum of content severity and social acceptability. This simplification could impact the generalizability of our models to real-world content moderation scenarios where more nuanced distinctions are crucial.

The annotation process presents limitations stemming from inherent subjectivity in erotic content interpretation. Despite achieving reasonable inter-annotator agreement (Krippendorff's Alpha of $0.716$ after outlier removal), the significant variation in individual annotator interpretations—particularly evident in the use of the \textit{ambiguous} category—suggests potential instability in ground truth labels.

From a technical perspective, our evaluation framework primarily focuses on classification accuracy without deeply examining model behavior on edge cases or adversarial examples. While our study demonstrates the superiority of Polish-specific models over multilingual alternatives, we acknowledge limitations in comparing across model architectures with significantly different parameter counts and training regimes.

Furthermore, a specific limitation pertaining to the evaluation phase is that the definitions of the labels were not explicitly included in the prompts provided to the models. We confirm this omission may have contributed to the lower performance of the models in comparison to the baseline.

These limitations underscore the need for future work addressing dataset diversity, annotation methodology refinement, and more robust evaluation frameworks for erotic content detection systems.

\section*{Acknowledgements}

The development of the PLLuM models used in the experiments was funded by the Polish Minister of Digital Affairs under a special purpose subsidy No. 1/WI/DBiI/2023: \textit{Responsible Development of the Open Large Language Model, PLLuM (Polish Large Language Model)}, aimed at supporting breakthrough technologies in the public and economic sectors, including an open, Polish-language intelligent assistant for public administration clients. The work on the dataset, however, was conducted independently and was not financed by this project.

\bibliography{custom}

\appendix

\clearpage
\section{Data Sources}
\label{appendix:data-sources}

All non-professional stories were scraped from publicly available websites including:
\begin{itemize}
\setlength{\itemsep}{0pt}
    \item \textit{opowiadaniaerotyczne-darmowo.com}
    \item \textit{sexopowiadania.pl}
    \item \textit{pornzone.com}
    \item \textit{anonserek.pl}
    \item \textit{opowi.pl} (categories: \textit{o życiu, różne, miłosne})
    \item \textit{opowiadania.pl}
    \item \textit{polki.pl}
\end{itemize}

\section{Annotation Guidelines Summary}\label{appendix:guidelines}
This annotation task aims to create a manually annotated dataset for erotic content detection, enabling the development of automated detection methods and content moderation systems. These systems can assist models in identifying undesired content that may be potentially harmful. Each annotator is provided with a set of samples to label exclusively with one of the five predefined categories. The annotation process is conducted independently, meaning that annotators do not have access to the labels or work of other team members. For the purpose of this task, \textit{a sample} is defined as a single sentence of varying length. The style and spelling of the sentences may vary significantly, and assessing the quality or correctness of the text is not part of the task.

Annotators are instructed to carefully read each sentence and assign one final label to it. Each sentence was presented in isolation, without additional context. They are also encouraged to pause or discontinue the task at any point if they feel uncomfortable with the content being evaluated.

Each sample should be exclusively assigned one of the following labels:
\begin{itemize}[nosep]
    \item Label \textit{e} (for \textit{erotic}): choose this label if the sentence describes sexual activities and desires, advanced flirting with evident erotic undertones, references to past sexual activities, or explicit sexual fantasies. 
    
    Please note that mere mentions of genital terms or publicly accepted romantic behaviors or neutral discussions of erotic or sexual topics (e.g., from the psychological perspective) are excluded from this category. 
    \item Label \textit{v} (for \textit{violence-related}): choose this label if the sentence depicts explicit sexual harassment, rape, lack of consent, or any other sort of non-consensual violence which concerns sexual intentions or activities. 
    
    Please note that a sentence can include violence which is not sex-related. Such instances are excluded from this category. Consensual BDSM behaviors are likely to fall under the label \textit{erotic}.
    \item Label \textit{u} (for \textit{socially unacceptable behaviors}): choose this label if the sentence describes sexual behaviors considered illegal, generally taboo, or violative of social norms, such as zoophilia, necrophilia, pedophilia, incest, and other sexual deviations. 
    
    Please note that if you classify a sample as both \textit{violence related} and featuring \textit{socially unacceptable behaviors}, the priority should be given to the latter. Such instances should be labeled as \textit{socially unacceptable behaviors}.
    \item Label \textit{a} (for \textit{ambiguous}): choose this label for sentences that may evoke some degree of erotic connotation but whose interpretation depends heavily on the context. If the sentence can reasonably be interpreted as non-erotic in a neutral context, it should be labeled as \textit{ambiguous}. 
    \item Label \textit{n} (for \textit{neutral}): choose this label if the sentence does not meet the criteria of the above mentioned sex-related categories. This typically includes non-sexual content on various topics, as well as general statements about human sexuality or intimate anatomy that do not carry erotic connotations.
\end{itemize}

\section{Prompt}

\begin{figure}[H]
\centering
\small
\begin{minipage}{\linewidth}
\begin{tcolorbox}[
    colback=gray!5,
    colframe=gray!15,
    boxrule=0.5pt,
    arc=2mm,
    outer arc=2mm,
    boxsep=3mm,
    left=2mm,
    right=2mm,
    top=2mm,
    bottom=2mm
]
\textbf{Prompt Structure:} \\[1mm]
You are a classifier, and your task is to analyze the text and assign one of the following labels:

\begin{enumerate}[leftmargin=*,itemsep=0pt,parsep=2pt]
    \item Neutral text
    \item Erotic text
    \item Ambiguous text
    \item Socially unacceptable behavior text
    \item Violence-related text
\end{enumerate}

\noindent Based on the text below, return the results in JSON format, which will contain the keys \texttt{"Text"} and \texttt{"Prediction"}. The value in the \texttt{"Prediction"} field should contain only the name of the class.

\noindent \textit{Input text:}
\end{tcolorbox}
\end{minipage}
\caption{Classification prompt template used for evaluating language models (translated into English).}
\label{fig:prompt-template-en}
\end{figure}

\newpage
\onecolumn
\section{Analysis of Type I and Type II errors}\label{appendix:errors}

\subsection{Full}

\begin{table*}[htbp] 
    \centering
    \scriptsize
    \label{tab:errorsfull}
    \begin{tabular*}{\textwidth}{@{\extracolsep{\fill}} c c c c c c c c c c c}
    \toprule
    \multirow{2}{*}{Model} & \multicolumn{2}{c}{Erotic} & \multicolumn{2}{c}{Ambiguous} & \multicolumn{2}{c}{Violence} & \multicolumn{2}{c}{Unacceptable} & \multicolumn{2}{c}{Neutral} \\
    \cmidrule(lr){2-3} \cmidrule(lr){4-5} \cmidrule(lr){6-7} \cmidrule(lr){8-9} \cmidrule(lr){10-11}
    & Type I & Type II & Type I & Type II & Type I & Type II & Type I & Type II & Type I & Type II \\
    \midrule
    HerBERT Base & 44& 42& 29& 39& 2& 4& 0& 2 & 58 & 46 \\
    HerBERT Large & 55 & 38 & 18 & 47 & 0 & 4 & 0 & 8 & 62 & 38 \\
    RoBERTa Base & 35 & 42 & 41& 40& 3& 2& 2& 1& 50& 46\\
    RoBERTa Large & 39& 41& 27& 41& 10& 2& 0& 3& 51& 40\\
    \midrule
    \textbf{0-shot}\\
    GPT-4o & 97 & 26& 14& 72& 32& 3& 228& 7& 1& 266\\
    C4AI Command-R& 165& 89& 99& 61& 18& 4& 329& 8& 9& 519\\
    Llama 3.1 70B-Instruct& 134& 19& 6& 72& 21& 3& 245&8 &8 & 312\\
    Mixtral 8x22B& 91& 73& 12& 72& 43& 2& 244& 5& 14& 252\\
    Bielik-11B-v2.3-Instruct& 260& 10& 6& 72& 29& 2& 144& 7& 0& 348\\
    PLLuM-Mistral-12B& 55& 108& 34& 70& 11& 1& 50& 8& 104& 64\\
    PLLuM-Mixtral-8x7B& 53& 85& 35& 69& 11& 1& 51& 7& 72& 60\\
    Llama-3.1-8B-PLLuM & 5& 238& 1& 72& 2& 3& 7& 8& 312& 6\\
    PLLuM-Mistral-12B (SFT)& 43& 34& 18& 72& 5& 0& 106& 8& 23& 80 \\
    Llama-3.1-8B-PLLuM (SFT)  & 47& 26& 39& 72& 1& 3& 76& 8& 27& 49\\
    \midrule
    \textbf{1-shot} \\
    GPT-4o & 99& 26& 14& 72& 33& 3& 255& 7& 1& 296 \\
    C4AI Command-R & 237& 30& 4& 71& 18& 4& 143& 8& 31& 320\\
    Llama 3.1 70B-Instruct & 83 & 50 & 15& 72& 21& 3& 250& 6& 8& 246\\
    Mixtral 8x22B & 87& 70& 19& 71& 26& 2& 178& 6& 30& 185\\
    Bielik-11B-v2.3-Instruct & 189& 14& 14& 71& 19& 3& 104& 6& 4& 236\\
    PLLuM-Mistral-12B & 52& 132& 50& 70& 7& 3& 54& 8& 119& 65\\
    PLLuM-Mixtral-8x7B & 66& 100& 63& 67& 8& 2& 90& 6& 64& 109\\
    Llama-3.1-8B-PLLuM & 53& 216& 20& 71& 36& 3& 44& 7& 260& 114\\
    PLLuM-Mistral-12B (SFT) & 35& 52& 34& 70& 7& 1& 102& 8& 23& 65\\
    Llama-3.1-8B-PLLuM (SFT) & 65& 24& 38& 72& 1& 3& 68& 8& 28& 62\\
    \midrule
    \textbf{5-shot} \\
    GPT-4o & 90& 34& 26& 72& 43& 1& 231& 8& 1& 276\\
    C4AI Command-R & 205& 105& 8& 72& 12& 3& 39& 8& 133& 209\\
    Llama 3.1 70B-Instruct & 85& 53& 15& 71& 24& 1& 272& 8& 6& 269\\
    Mixtral 8x22B & 142& 48& 16& 72& 33& 2& 154& 5& 17& 233\\
    Bielik-11B-v2.3-Instruct & 172& 19& 7& 72& 24& 2& 138& 6& 1& 239\\
    PLLuM-Mistral-12B & 55& 108& 34& 70& 11& 1& 50& 8& 104& 64\\
    PLLuM-Mixtral-8x7B & 53& 85& 35& 69& 11& 1& 51& 7& 72& 60\\
    Llama-3.1-8B-PLLuM & 30& 288& 16& 72& 20& 3& 10& 8& 277& 51\\
    PLLuM-Mistral-12B (SFT) & 43& 34& 18& 72& 5& 0& 106& 8&23 &80 \\
    Llama-3.1-8B-PLLuM (SFT) & 75& 28& 42& 72& 3& 3& 73& 8& 31& 86\\
    \bottomrule
    \end{tabular*}
    \caption{Detailed Type I and Type II error counts for various models for \textit{Full} dataset type.}
\end{table*}

\newpage
\subsection{Extended}

\begin{table*}[htbp] 
    \centering
    \scriptsize
    \label{tab:errorsextended}
    \begin{tabular*}{\textwidth}{@{\extracolsep{\fill}} c c c c c c c c c}
    \toprule
    \multirow{2}{*}{Model} & \multicolumn{2}{c}{Erotic} & \multicolumn{2}{c}{Ambiguous} & \multicolumn{2}{c}{Violence + Unacceptable} & \multicolumn{2}{c}{Neutral} \\
    \cmidrule(lr){2-3} \cmidrule(lr){4-5} \cmidrule(lr){6-7} \cmidrule(lr){8-9}
    & Type I & Type II & Type I & Type II & Type I & Type II & Type I & Type II \\
    \midrule
    HerBERT Base & 41& 45& 34& 40& 3& 7& 58& 44 \\
    HerBERT Large &  33& 58& 41& 33& 6& 3& 54&40\\
    RoBERTa Base & 45& 49& 38& 42& 8& 2& 48&51\\
    RoBERTa Large & 42& 37& 30& 43& 4& 4& 47&44\\
    \midrule
    \textbf{0-shot}\\
    GPT-4o & 95& 33& 240& 34& 42& 10& 2& 262 \\
    C4AI Command-R& 75& 216& 26& 71& 9& 11& 265& 76 \\
    Llama 3.1 70B-Instruct& 75& 101& 220& 34& 22& 9& 59& 232 \\
    Mixtral 8x22B& 52& 99& 225& 36& 39& 9& 39& 174 \\
    Bielik-11B-v2.3-Instruct& 159& 42& 66& 63& 19& 10& 44& 160 \\
    PLLuM-Mistral-12B& 78& 91& 40& 56& 100& 5& 55& 104 \\
    PLLuM-Mixtral-8x7B& 116& 44& 62& 39& 43& 9& 26& 112 \\
    Llama-3.1-8B-PLLuM & 7& 256& 3& 72& 2& 12& 334& 6 \\
    PLLuM-Mistral-12B (SFT)& 31& 47& 68& 12& 16& 12& 42& 37 \\
    Llama-3.1-8B-PLLuM (SFT)  & 58& 24& 73& 36& 10& 12& 28& 49 \\
    \midrule
    \textbf{1-shot} \\
    GPT-4o & 117& 28& 241& 37& 44& 10& 1& 293 \\
    C4AI Command-R & 221& 32& 149& 70& 21& 10& 35& 310 \\
    Llama 3.1 70B-Instruct & 82& 59& 201& 32& 42& 7& 12& 239 \\
    Mixtral 8x22B & 87& 66& 166& 43& 80& 7& 27& 185 \\
    Bielik-11B-v2.3-Instruct & 194& 27& 95& 64& 35& 9& 6& 213 \\
    PLLuM-Mistral-12B & 56& 105& 39& 65& 56& 5& 93& 55 \\
    PLLuM-Mixtral-8x7B & 67& 89& 96& 42& 56& 7& 56& 102 \\
    Llama-3.1-8B-PLLuM & 60& 208& 65& 71& 32& 11& 255& 120 \\
    PLLuM-Mistral-12B (SFT) & 43& 40& 122& 29& 21& 11& 29& 61 \\
    Llama-3.1-8B-PLLuM (SFT) & 56& 105& 39& 65& 56& 5& 93& 55 \\
    \midrule
    \textbf{5-shot} \\
    GPT-4o & 88& 44& 256& 35& 62& 9& 2& 281 \\
    C4AI Command-R & 193& 110& 70& 33&20 & 8& 136&  225\\
    Llama 3.1 70B-Instruct & 85& 52& 272& 31& 48& 7& 8& 258 \\
    Mixtral 8x22B & 130& 110& 70& 33& 20& 8& 136& 225 \\
    Bielik-11B-v2.3-Instruct & 153& 23& 163& 53& 27& 9& 2& 239 \\
    PLLuM-Mistral-12B & 64& 101& 50&42 & 38& 10& 100& 68 \\
    PLLuM-Mitral-8x7B & 66& 89& 66& 41& 44& 10& 76& 76 \\
    Llama-3.1-8B-PLLuM & 19& 237& 24& 68& 25& 10& 296& 36 \\
    PLLuM-Mistral-12B (SFT) & 60& 28& 87& 34& 26& 11& 19& 80 \\
    Llama-3.1-8B-PLLuM (SFT) & 72& 26& 70& 44& 22& 10& 27& 80 \\
    \bottomrule
    \end{tabular*}
    \caption{Detailed Type I and Type II error counts for various models for \textit{Extended} dataset type.}
\end{table*}

\newpage
\subsection{Core}

\begin{table*}[htbp] 
    \centering
    \scriptsize
        \label{tab:errorscore}
    \begin{tabular*}{\textwidth}{@{\extracolsep{\fill}} c c c c c c c}
    \toprule
    \multirow{2}{*}{Model} & \multicolumn{2}{c}{Erotic} & \multicolumn{2}{c}{Ambiguous} & \multicolumn{2}{c}{Neutral} \\
    \cmidrule(lr){2-3} \cmidrule(lr){4-5} \cmidrule(lr){6-7}
    & Type I & Type II & Type I & Type II & Type I & Type II  \\
    \midrule
    HerBERT Base & 53& 40& 47& 55& 56& 54 \\
    HerBERT Large & 48& 36& 48& 45& 44& 59 \\
    RoBERTa Base & 37& 46& 46& 46& 62& 51 \\
    RoBERTa Large & 43& 40& 49& 43& 51& 55 \\
    \midrule
    \textbf{0-shot}\\
    GPT-4o& 74& 25& 301& 23& 1& 279 \\
    C4AI Command-R& 68& 238& 45& 68& 284& 88 \\
    Llama 3.1 70B-Instruct& 57& 96& 320& 30& 48& 257 \\
    Mixtral 8x22B& 58& 90& 241& 30& 35& 172 \\
    Bielik-11B-v2.3-Instruct& 142& 21& 109& 59& 20& 178 \\
    PLLuM-Mistral-12B& 87& 39& 98& 52& 36& 110 \\
    PLLuM-Mixtral-8x7B& 106& 32& 125& 49& 17& 144 \\
    Llama-3.1-8B-PLLuM & 10& 239& 8& 72& 305& 12 \\
    PLLuM-Mistral-12B (SFT)& 22& 56& 67& 34& 60& 21 \\
    Llama-3.1-8B-PLLuM (SFT)& 32& 30& 79& 39& 38& 47 \\
    \midrule
    \textbf{1-shot} \\
    GPT-4o& 95& 21& 327& 28& 0& 329 \\
    C4AI Command-R& 206& 37& 232& 62& 36& 375 \\
    Llama 3.1 70B-Instruct& 62& 55& 243& 22& 11& 239 \\
    Mixtral 8x22B& 79& 61& 229& 31& 18& 193 \\
    Bielik-11B-v2.3-Instruct& 140& 19& 181& 52& 2& 232 \\
    PLLuM-Mistral-12B& 89& 57& 86& 60& 67& 113 \\
    PLLuM-Mixtral-8x7B& 70& 55& 169& 39& 24& 136 \\
    Llama-3.1-8B-PLLuM & 72& 197& 92& 69& 232& 127 \\
    PLLuM-Mistral-12B (SFT)& 34& 43& 70& 33& 50& 39 \\
    Llama-3.1-8B-PLLuM (SFT)& 38& 24& 96& 31& 34& 58 \\
    \midrule
    \textbf{5-shot} \\
    GPT-4o& 88& 29& 303& 29& 0& 290 \\
    C4AI Command-R& 185& 117& 129& 71& 138& 236 \\
    Llama 3.1 70B-Instruct& 67& 57& 269& 19& 5& 265 \\
    Mixtral 8x22B& 105& 37& 253& 41& 16& 243 \\
    Bielik-11B-v2.3-Instruct& 106& 31& 240& 44& 1& 244 \\
    PLLuM-Mistral-12B& 102& 54& 62& 62& 67& 97 \\
    PLLuM-Mixtral-8x7B& 45& 75& 165& 38& 40& 103 \\
    Llama-3.1-8B-PLLuM & 24& 247& 43& 69& 290& 38 \\
    PLLuM-Mistral-12B (SFT)& 56& 28& 63& 38& 31& 50 \\
    Llama-3.1-8B-PLLuM (SFT)& 45& 24& 64& 40& 46& 49 \\
    \bottomrule
    \end{tabular*}
    \caption{Detailed Type I and Type II error counts for various models for \textit{Core} dataset type.}
\end{table*}

\newpage
\subsection{Basic}
\begin{table*}[htbp] 
    \centering
    \scriptsize
    \label{tab:errorsbasic}
    \begin{tabular*}{\textwidth}{@{\extracolsep{\fill}} c c c c c}
    \toprule
    \multirow{2}{*}{Model} & \multicolumn{2}{c}{Erotic} & \multicolumn{2}{c}{Neutral} \\
    \cmidrule(lr){2-3} \cmidrule(lr){4-5}
    & Type I & Type II & Type I & Type II \\
    \midrule
    HerBERT Base & 34& 34& 34& 34 \\
    HerBERT Large & 31& 28& 28& 31 \\
    RoBERTa Base & 31& 23& 23& 31 \\
    RoBERTa Large & 25& 29& 29& 25 \\
    \midrule
    \textbf{0-shot}\\
    GPT-4o& 115& 1& 1& 115\\
    C4AI Command-R& 76& 191& 191& 76 \\
    Llama 3.1 70B-Instruct& 152& 19& 19& 152 \\
    Mixtral 8x22B& 95& 33& 33& 95 \\
    Bielik-11B-v2.3-Instruct& 140& 18& 18& 140 \\
    PLLuM-Mistral-12B& 87& 20& 20& 87 \\
    PLLuM-Mixtral-8x7B& 127& 4& 4& 127 \\
    Llama-3.1-8B-PLLuM & 6& 246& 246& 6 \\
    PLLuM-Mistral-12B (SFT)& 17& 34& 34& 17 \\
    Llama-3.1-8B-PLLuM (SFT)& 31& 20& 20& 31 \\
    \midrule
    \textbf{1-shot} \\
    GPT-4o& 132& 1& 1& 132 \\
    C4AI Command-R& 263& 34& 34& 263 \\
    Llama 3.1 70B-Instruct& 153& 10& 10& 153 \\
    Mixtral 8x22B& 134& 18& 18& 134 \\
    Bielik-11B-v2.3-Instruct& 149& 1& 1& 149 \\
    PLLuM-Mistral-12B& 127& 35& 35& 127 \\
    PLLuM-Mixtral-8x7B& 79& 19& 19& 79 \\
    Llama-3.1-8B-PLLuM & 76& 156& 156& 76 \\
    PLLuM-Mistral-12B (SFT)& 40& 47& 40&  \\
    Llama-3.1-8B-PLLuM (SFT)& 127& 35& 35& 127 \\
    \midrule
    \textbf{5-shot} \\
    GPT-4o& 113& 0& 0& 113 \\
    C4AI Command-R& 199& 96& 96& 199 \\
    Llama 3.1 70B-Instruct& 159& 4& 4& 159 \\
    Mixtral 8x22B& 167& 10& 10& 167 \\
    Bielik-11B-v2.3-Instruct& 135& 3& 3& 135 \\
    PLLuM-Mistral-12B& 111& 46& 46& 111 \\
    PLLuM-Mixtral-8x7B& 44& 32& 32& 44 \\
    Llama-3.1-8B-PLLuM & 28& 200& 200& 28 \\
    PLLuM-Mistral-12B (SFT)& 73& 15& 15& 73 \\
    Llama-3.1-8B-PLLuM (SFT)& 43& 17& 17& 43 \\
    \bottomrule
    \end{tabular*}
    \caption{Detailed Type I and Type II error counts for various models for \textit{Basic} dataset type.}
\end{table*}

\twocolumn
\section{Computational Details}
The details provided below correspond to the \textbf{Full} version of the dataset.

The encoder-based models were trained for approximately 5 hours and 10 hours, respectively, for the Base and Large versions, using A100 (40GB) GPU with early stopping. The batch size was set to 8 for both model variants. 
Training of the PLLuM family models on our dataset took approximately 3 hours for the Llama-3.1-8B-PLLuM (SFT) model and 4 hours for the PLLuM-Mistral-12B (SFT) model. These models were trained on a distributed setup consisting of 2 nodes, each equipped with 4 H100 (96GB) GPUs, ensuring parallel processing across GPUs for enhanced training efficiency.






\end{document}